\def\year{2021}\relax
\documentclass[letterpaper]{article} 
\usepackage{aaai21}  
\usepackage{times}  
\usepackage{helvet} 
\usepackage{courier}  
\usepackage[hyphens]{url}  
\usepackage{graphicx} 
\usepackage{natbib}  
\usepackage{caption} 
\usepackage{amsmath,amssymb}
\usepackage{booktabs}
\usepackage{epsfig}
\usepackage{multirow}
\usepackage{graphics}
\usepackage{subcaption}
\title{Visual Comfort Aware-Reinforcement Learning \\ for Depth Adjustment of Stereoscopic 3D Images}
\author{
	Hak Gu Kim\textsuperscript{\rm 1,2}\thanks{Work done as a part of the research project in KAIST},
	Minho Park\textsuperscript{\rm 1},
	Sangmin Lee\textsuperscript{\rm 1},
	Seongyeop Kim\textsuperscript{\rm 1},
	Yong Man Ro\textsuperscript{\rm 1}\thanks{Corresponding author (ymro@kaist.ac.kr)}
	\\
}
\affiliations{
	\textsuperscript{\rm 1}Image and Video Systems Lab., KAIST, Korea\\
	\textsuperscript{\rm 2}School of Computer and Communication Sciences, EPFL, Switzerland
	\\
	hakgu.kim@epfl.ch, \{roger618, sangmin.lee, seongyeop, ymro\}@kaist.ac.kr
}

\begin{document}

\maketitle

\begin{abstract}
Depth adjustment aims to enhance the visual experience of stereoscopic 3D (S3D) images, which accompanied with improving visual comfort and depth perception. For a human expert, the depth adjustment procedure is a sequence of iterative decision making. The human expert iteratively adjusts the depth until he is satisfied with the both levels of visual comfort and the perceived depth. In this work, we present a novel deep reinforcement learning (DRL)-based approach for depth adjustment named \textit{VCA-RL} (\textit{Visual Comfort Aware Reinforcement Learning}) to explicitly model human sequential decision making in depth editing operations. We formulate the depth adjustment process as a Markov decision process where actions are defined as camera movement operations to control the distance between the left and right cameras. Our agent is trained based on the guidance of an objective visual comfort assessment metric to learn the optimal sequence of camera movement actions in terms of perceptual aspects in stereoscopic viewing. With extensive experiments and user studies, we show the effectiveness of our VCA-RL model on three different S3D databases.
\end{abstract}


\begin{figure*}
\begin{center}
\includegraphics[width=0.8\linewidth] {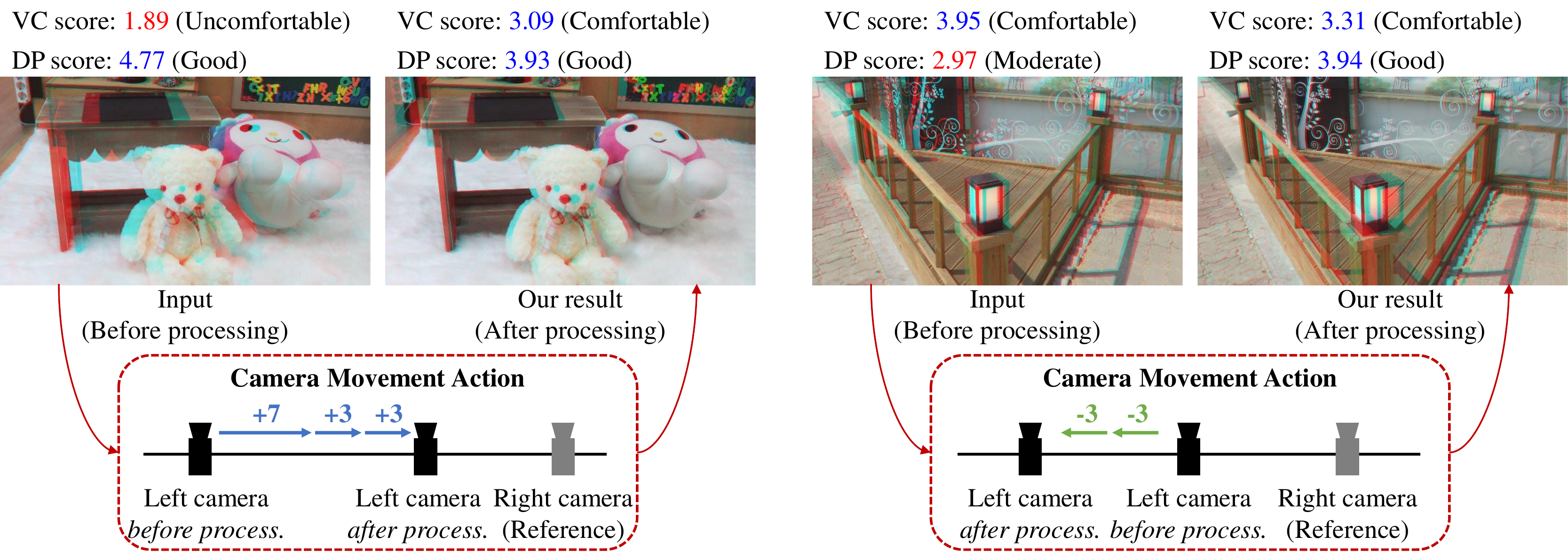}
\end{center}
   \caption{The intuition of the proposed VCA-RL model. We train our agent guided by visual comfort assessment metric. Similar to human professionals, the agent sequentially determines the camera movement action. By applying the action, we can obtain the visually comfortable stereoscopic image with sufficient depth. Note that the VC score is a visual comfort score and the DP score is a depth perception score.}
\label{fig:1}
\end{figure*}


\section{Introduction}
With the concerns on the viewing safety in stereoscopic 3D (S3D) displays, depth adjustment has increasingly gained importance for improving visual experience of stereoscopic images such as visual comfort and depth perception \cite{1,2,3}. For a proper viewing experience of S3D contents, highly skilled professionals (e.g., stereographers) carefully control the camera parameters such as a camera baseline using professional depth editing tools. It requires not only expertise in stereoscopy, but also a lot of time and effort \cite{3}. Therefore, it is essential to develop an automatic depth adjustment method. 

Previous studies have proposed various depth adjustment methods to improve visual comfort by shifting the zero disparity plane (ZDP) or scaling the disparity range of a scene. However, a common shortcoming of these existing works is that they edited the given disparities \textit{in a direct way} without considering the perceptual effects of the changed depth and visual comfort levels. In addition, they mainly focused on the visual comfort improvement rather than the depth perception. There is a trade-off between the visual comfort improvement and the perceived depth degradation in the depth adjustment process. That is why human experts carefully and iteratively manipulate depths, \textit{not in a direct way}.

For a human expert, the depth adjustment procedure is a sequence of iterative decision making for S3D contents. A human expert iteratively conducts depth editing operations until the levels of visual comfort and the perceived depth fit what he has in mind. Adjusting depths to the optimal is complex operations that need to consider the perceptual aspect as well as spatial distortions. The perceptual effect should be taken into account to prevent undesirable perceptual side-effects such as excessive visual discomfort or unnoticeable depth in stereoscopic viewing.

Inspired by human expert's sequential decision making which benefits depth editing, we propose a novel depth adjustment framework by combining the knowledge of human binocular perception and deep reinforcement learning named \textit{VCA-RL} (\textit{Visual Comfort Aware Reinforcement Learning}). Despite recent advances in deep learning-based S3D applications (e.g., visual comfort assessment \cite{4,5} and stereo matching \cite{6,7,8,9}), it is hard to extend these approaches to the depth adjustment task due to the complex non-linear operation, subjective nature of human visual system, and the lack of expensive pairs of input and processed S3D contents. In this paper, we firstly formulate the depth adjustment process as a Markov decision process to control the distance between the left and right cameras (i.e., stereo baseline) in a sequential way. By iteratively adjusting the stereo baseline via camera movement actions, the range of disparity could be carefully controlled to achieve satisfying visual experiences (see Fig.~\ref{fig:1}).	In particular, to find a proper 3D visual satisfaction comprising visual comfort and perceived depth, we design a novel visual comfort aware reward function based on the guidance of objective visual comfort assessment metric.	Based on the visual comfort aware reward, our agent can learn the optimal sequence of camera movement actions preserving both visual comfort and perceived depth in stereoscopic viewing. 

The contributions of this work are summarized as follows.
\begin{itemize}
\item Inspired by human expert's iterative decision making in depth editing, we firstly design a depth adjustment agent using reinforcement learning that learns iterative depth adjustment process. By sequentially adjusting the stereo baseline in the world coordinates, our model can find an optimal trade-off between visual comfort improvement and the perceived depth degradation.
\item We propose a novel visual comfort aware reward function. By learning the reward based on the predicted visual comfort scores of the stereoscopic image at each step in training, our VCA-RL model can automatically decide for itself whether to ameliorate the visual comfort or improve the depth perception at each step in testing. 
\item With extensive experiments and subjective evaluations, we demonstrate the effectiveness and the superiority of our VCA-RL model for improving visual experiences of stereoscopic images  on various S3D databases.
\end{itemize}


\begin{figure*}
\begin{center}
\includegraphics[width=0.88\linewidth] {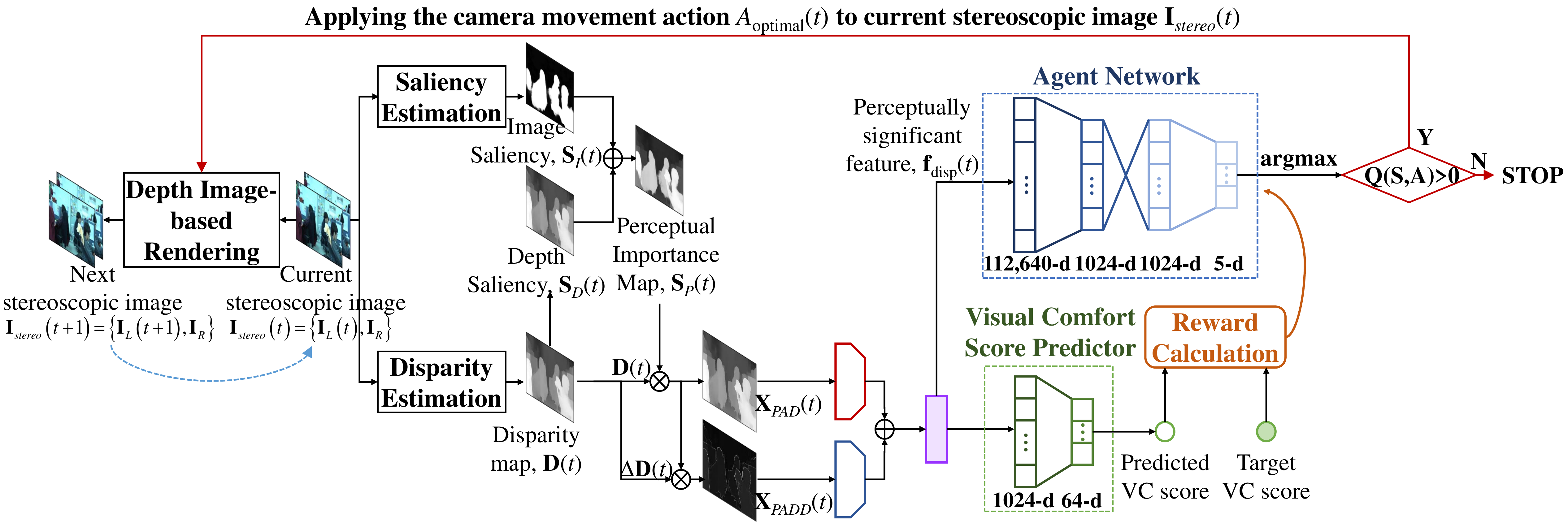}
\end{center}
   \caption{The illustration of the proposed VCA-RL framework for depth adjustment. At first, the perceptual importance map $\textbf{S}_P$ is estimated from $\textbf{I}_L$ and $\textbf{D}_L$. Based on that, we encode $\textbf{f}_{disp}$. Then, our agent estimates the action value $\mathcal{Q}(\mathcal{S}(t), \mathcal{A})$ that maximizes the visual comfort aware reward. The selected camera movement action is applied to the input for depth adjustment. This process is iteratively carried out until the visual comfort of stereoscopic image falls into the comfortable range.}
\label{fig:2}

\end{figure*}


\section{Related Work}
Depth adjustment mainly aims at improving visual comfort of stereoscopic images while preserving the perceived depth. The visual discomfort is highly related with the disparity/depth characteristics of stereoscopic images (e.g., disparity magnitude \cite{10,11} and disparity difference \cite{12,13}). To deal with that, there are two main approaches to for depth adjustment, which are disparity shifting \cite{14,15,16} and disparity scaling \cite{17,18,19,20,21,22,23}.

Previous works have proposed various disparity shifting methods to reduce visual discomfort by simply moving the ZDP of the original scene while maintaining the range of disparity. Shao \textit{et al}. \cite{15} proposed a disparity shifting method considering spatial frequency, disparity response, and visual attention to mitigate visual discomfort in stereoscopic viewing. Recently, Ying \textit{et al}. \cite{16} proposed a viewing distance-based nonlinear shifting (VDNS) approach to improve visual comfort and perceived depth quality. The disparity shifting methods are effective to reduce excessive screen disparity. They can mitigate the accommodation-vergence (AV) conflict \cite{24,25} with a low computational cost. However, they cannot reduce visual fatigue of stereoscopic images with disparity range exceeding visual comfortable zone (i.e., $\pm1^{\circ}$ angular disparity \cite{2,3})\cite{19}. In the proposed method, by explicitly adjusting the distance of stereo cameras, the overall depth range can be edited to fit a visual comfortable zone based on the guidance of the objective visual comfort assessment metric.

The disparity scaling methods have been proposed that linearly or nonlinearly adjusted the disparity range of stereoscopic images into the visual comfortable zone. Lang \textit{et al}. \cite{17} proposed a nonlinear disparity mapping based on visual importance of scene elements. Sohn \textit{et al}.\cite{18} proposed a disparity remapping method combining global and local disparity range adjustments. Jung \textit{et al}. \cite{20} proposed a visual comfort improvement method that adaptively adjusted the depth range considering saliency information. Shao \textit{et al}. \cite{23} developed an optimization-based approach that conducted layer-dependent depth range adjustment considering both visual comfort and depth sensation. However, the disparity scaling way can lead to decrease the relative distance between objects because the scene is compressed intentionally. It can also reduce the senses of the perceived depth and realism. In addition, by increasing the viewing distance between eyes and the scene, the sense of presence can be weakened \cite{16}. On the other hand, we do not explicitly change the disparities in the image domain. In the proposed method, we can preserve the geometric proportion of objects and the relative distance between objects in 3D space because we progressively adjusted the distance between left and right cameras in the world coordinate.


\section{Visual Comfort Aware-Reinforcement Learning for Depth Adjustment}

To imitate the expert's decision making process, we formulate the depth adjustment as a problem of finding an optimal sequence of camera movement action $\mathcal{A}$. We adjust depth of given left image $\textbf{I}_{L}$ and right image $\textbf{I}_{R}$ by iteratively applying the camera movement action $\mathcal{A}$. The visual comfort score ${s}_{VC}(t)$ at a step $t$ is estimated from the disparity map $\textbf{D}(t)$ and the perceptual importance map $\textbf{S}_P(t)$. Based on a sequence of the predicted comfort score $\hat{s}_{VC}$ at each step, our agent determines a camera movement action $\mathcal{A}(t)$ for depth adjustment under the policy ${\Omega}_{\theta}$. Therefore, our goal is to find an optimal sequence for depth adjustment $\mathcal{T}\left \{ \mathcal{A}_{optimal}(t)\subset \mathcal{A}\right \}$ that presents visually comfortable and sufficient depth in stereoscopic viewing.

Fig.~\ref{fig:2} shows the overall process of the proposed VCA-RL framework for depth adjustment. At first, the disparity map $\textbf{D}(t)$ and the perceptual importance map $\textbf{S}_P(t)$ are estimated from given stereoscopic image at a step $t$, $\textbf{I}_{stereo}=[\textbf{I}_{L}(t), \textbf{I}_{R} ]$ ($\textbf{I}_{R}$ is used as a reference in our study). By considering the disparity and human attention information, the perceptually significant disparity feature $\textbf{f}_{disp}(t)$ is encoded. Then, $\textbf{f}_{disp}(t)$ is forwarded to visual comfort score predictor (VC score predictor) to evaluate the degree of visual comfort, $\hat{s}_{VC}(t)$. $\textbf{f}_{disp}(t)$ is also forwarded to our agent network to estimate the action value $\mathcal{Q}(\mathcal{S}(t), \mathcal{A})$, which is the expected sum of future reward $\mathcal{R}$. The state $\mathcal{S}(t)$ is a combination of $\textbf{f}_{disp}(t)$ and $\hat{s}_{VC}(t)$, $\mathcal{S}\left ( t \right )=\left \{ \mathbf{f}_{disp}(t), \hat{s}_{VC}(t) \right \}$. The agent then approximates the action value $\mathcal{Q}(\mathcal{S}(t), \mathcal{A})$ and chooses the best action $\mathcal{A}_{optimal}(t)$ maximizing the action value $\mathcal{Q}(\mathcal{S}(t), \mathcal{A})$. Finally, by applying the best action at a step $t$, $\mathcal{A}_{optimal}(t)$ to input $\textbf{I}_{stereo}(t)$, the stereoscopic image at next step $t+1$, $\textbf{I}_{stereo}(t+1)$, is obtained via depth image based rendering (DIBR) with the updated stereo baseline. The agent repeats this process and stops when all estimated action values are negative.

\subsection{Camera Movement Action}
To adjust the range of depth, the action $\mathcal{A}$ is composed by the camera movements. By explicitly increasing or decreasing the distance between stereo cameras (i.e., stereo baseline) in the world coordinates, we can manipulate the depth of stereoscopic image while preserving relative distance between objects and their geometric proportions. The camera movement action at a step $t$, $\mathcal{A}(t)$, is only applied to the left camera. The right camera is fixed (i.e., reference). In this work, we define 5 camera movement actions, $\mathcal{A}=\left \{-7, -3, +3, +7, 0 \right \}$, to shift the position of the left camera on the stereo camera baseline. The sign indicates the direction the left camera moves ($'-'$ for left side and $'+'$ for right side). The values mean the distance (unit: $mm$) that the camera moves at each step. They are determined in consideration of the distance between the pupils of eyes (i.e., interpupillary distance $\simeq 63 mm$). The zero means the termination of iterative depth adjustment operations.

\subsection{Perceptually Significant Disparity Feature}
It is well known that the disparity magnitude, which is related with the absolute screen disparity, is a critical factor affecting visual discomfort due to binocular fusion limit (i.e., Panum’s fusional area \cite{26}) \cite{10,11}. The disparity gradient, which is the disparity difference between nearby objects (i.e., differential disparity), reflects on visual discomfort as well \cite{12,13}. Based on these characteristics, we employ the perception-weighted absolute disparity map (PAD) and the perception-weighted differential disparity map (PADD) to encode perceptually significant disparity feature as in \cite{4,27}. For this purpose, We first generate the perceptual importance map $\textbf{S}_{P}$ using both image saliency $\textbf{S}_{I}$ and disparity saliency $\textbf{S}_{D}$. For $\textbf{S}_{I}$, we employ a recent deep learning-based saliency estimation \cite{28}. Note that the saliency values range from 0 (least saliency) to 1 (most saliency). For $\textbf{S}_{D}$, we assume that the foreground objects usually attract more human attention compared with backgrounds in a scene \cite{4,27}. $\textbf{S}_{D}$ is generated by mapping the minimum and maximum disparity values in $\textbf{D}_{L}$ to 0 and 1, respectively. In this study, a hierarchical deep stereo matching (HSM) \cite{6} is used for disparity estimation. Finally, the perceptual importance map $\textbf{S}_{P}$ is computed (see Fig. S1 in our supplementary file), which can be written as

\begin{equation}
	\label{eq:1}
		\textbf{S}_{P} = {w}_{I}\textbf{S}_{I} + {w}_{D}\textbf{S}_{D}
\end{equation}
where we set ${w}_{I}={w}_{D}=0.5$ in our experiment.

Then, we obtain the PAD, $\textbf{X}_{PAD}=\textbf{S}_{P}\otimes|\textbf{D}|$, and PADD, $\textbf{X}_{PADD}=\textbf{S}_{P}\otimes|\Delta \textbf{D}|$ where $\otimes$ indicates element-wise multiplication. We use them as input of our perceptual feature extractor. To encode the perceptually significant disparity feature $\textbf{f}_{disp}\in \mathbb{R}^{11\times 10\times 1024}$ capturing the visual comfort level of stereoscopic images, we employ a deep convolutional neural network (DCNN) based on VGG-16 \cite{29,4}. The disparity feature is trained by $f\left ( \cdot \right )$ and regressed to visual comfort score by $p\left ( \cdot \right )$. During this training, by minimizing the loss for visual comfort prediction $L_{VC}$, the perceptually significant disparity feature $\textbf{f}_{disp}$ is encoded.
\begin{equation}
	\label{eq:2}
		L_{VC}=\frac{1}{N}\sum_{i=1}^{N}\left \| p\left ( \textbf{f}_{disp}^{i} \right ) - s_{VC}^{i}  \right \|^2
\end{equation}
where $p\left ( \textbf{f}_{disp}^{i} \right )$ is the predicted comfort score for $i$-th stereo image, (i.e., $\hat{s}_{VC}^{i}$) and $s_{VC}^{i}$ is the corresponding ground-truth comfort score. $N$ is the number of training dataset.

\subsection{Visual Comfort Aware Reward}
To make our agent determine an optimal camera movement action sequence in terms of viewing experience of S3D contents, we design a novel visual comfort aware reward function using the objective visual comfort assessment metric (i.e., visual comfort score) for stereoscopic images.

The visual comfort score can be divided into 5-scale, which are 1: extremely uncomfortable, 2: uncomfortable, 3: comfortable, 4: moderately comfortable, and 5: Very comfortable \cite{23}. We reasonably assume that $s_{VC}^{T}=3$ (comfortable) is the target comfort level while maintaining the sufficient depth. This is because the level of visual comfort is inversely related to the level of the perceived depth in stereoscopic viewing.

Our goal is to find the optimal sequence of camera movement actions for depth adjustment $\mathcal{T}\left \{ \mathcal{A}_{optimal}(t)\subset \mathcal{A}\right \}$ that minimizes the difference between the predicted comfort score of a given stereoscopic image and the target comfort score. The process can be regarded as a Markov decision process. In a Markov decision process, the state $\mathcal{S}$ is a combination of the $\textbf{f}_{disp}$ and $\hat{s}_{VC}$. The action space is a set of our camera movement operations $\mathcal{A}$. Finally, inspired by \cite{31}, our visual comfort aware reward $\mathcal{R}(t)$ can be defined as
\begin{equation}
	\label{eq:3}
		\mathcal{R}(t)=sign\left ( - \left | s_{VC}^{T} - \hat{s}_{VC}(t+1) \right | + \left | s_{VC}^{T} - \hat{s}_{VC}(t) \right |\right )
\end{equation}
where $sign(\cdot)$ is a sign function. In our study, the sign function is used to limit the variation of the difference values and make model training stable \cite{30}.

In our VCA-RL model, if the distance from the target comfort score is lower than 0.3, the positive reward is given to our agent. If the distance from target comfort score is higher than 0.3, our agent will receive a negative reward as a penalty for the action \cite{32}.
\begin{equation}
	\label{eq:4}
		\mathcal{R}(\mathcal{S}(t), \mathcal{A}(t))=\begin{cases}
+\eta,  & \text{ if } \left |s_{VC}^{T}- \hat{s}_{VC} \right | < 0.3\\ 
-\eta,  & \text{ otherwise} 
\end{cases}
\end{equation}
where $\eta$ set to 0.3 in our experience.

Our reward function is to adjust stereo baseline so that the comfort score at $t+1$ is closer to the target score than before. Otherwise, the action is penalized. Through the proposed reward function, the agent can learn the rules about which action should be chosen as $\mathcal{A}_{optimal}(t)$.

\subsection{Agent for Depth Adjustment}
Our agent network consists of 4 fully connected layers for action value estimation. The perceptually significant disparity feature $\textbf{f}_{disp}(t)$ is fed to our agent network. The agent estimates the action value $\mathcal{Q}(\mathcal{S}(t),\mathcal{A})$ with Q-learning. It can be defined as an expected sum of future visual comfort aware rewards \cite{33}. The Q-learning iteratively updates the action-selection policy $\mathrm{\Omega}_{\Theta}$ using the Bellman equation, which can be written as

\begin{equation}
	\label{eq:5}
		\begin{split}
		\mathcal{Q}(\mathcal{S}(t), \mathcal{A}) &= E\left [ \mathcal{R}(t)+\gamma \mathcal{R}(t+1)+\gamma^2 \mathcal{R}(t+2) +\cdots \right ]\\ 
		&\simeq \mathcal{R}(t)+\gamma\max_{\mathcal{A'}}\mathcal{Q}(\mathcal{S}(t),\mathcal{A})
		\end{split}
\end{equation}
where $\gamma$ is a discount factor and set to 0.9 \cite{32}. We train the agent to estimate the action value $\mathcal{Q}(\mathcal{S}(t), \mathcal{A}(t))$ and choose an optimal action $\mathcal{A}_{optimal}(t)$ that maximizes $\mathcal{Q}(\mathcal{S}(t), \mathcal{A})$.

To train the agent network, we use an $\epsilon$-greedy algorithm. By the $\epsilon$-greedy algorithm, the policy $\mathrm{\Omega}_{\Theta}$ is determined during training. The $\epsilon$-greedy algorithm randomly samples actions with a probability of $\epsilon$ and takes the actions with the highest reward in a greedy way with a probability of 1-$\epsilon$. In the test stage, the policy is determined with $\epsilon=0$, i.e. the highest expected reward is always chosen. The process is repeated until all expected rewards are negative.

After our agent chooses $\mathcal{A}_{optimal}(t)$, the action is applied to $\textbf{I}_{stereo}$ to edit its depth range. By using DIBR process, we can synthesize a new left image $\textbf{I}_L(t+1)$ at a new left camera position moved by the selected camera movement action. In this work, the disocclusions are very small in $\textbf{I}_L(t+1)$ because the camera is progressively moved to the optimal position. In our experiment, the disoccluded regions in $\textbf{I}_L(t+1)$ are filled with the hole filling method considering binocular symmetry \cite{38}. As noted, this study focuses on formulating the depth adjustment framework as a sequential decision making process like a human expert, rather than the development of a new image-based rendering method.


\begin{figure*}[t!]
    \centering
    
    \begin{subfigure}{0.85\textwidth}
    	\centering
        	\setlength{\abovecaptionskip}{1.0 pt}
            \includegraphics[width=1.0\textwidth,keepaspectratio]{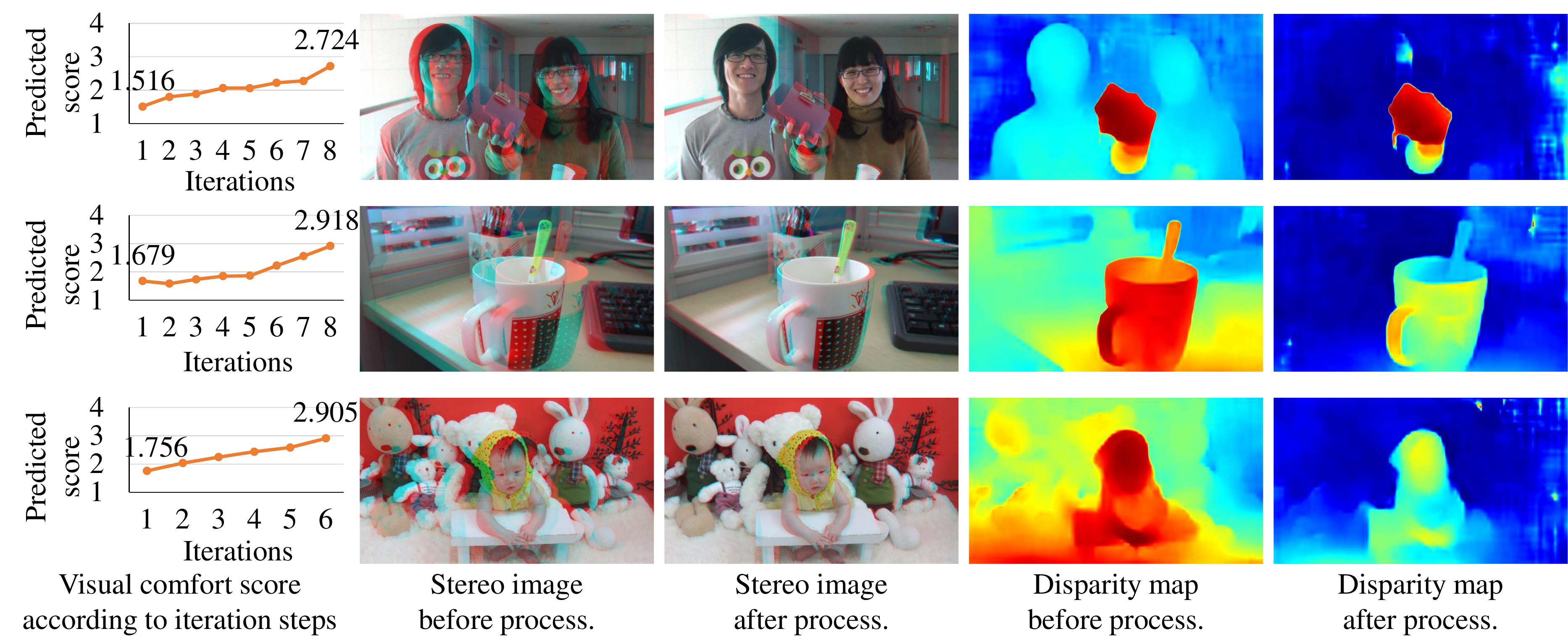}
            \caption{}
			\label{fig3:a}
    \end{subfigure}

	\begin{subfigure}{0.85\textwidth}
		\centering
        	\setlength{\abovecaptionskip}{1.0 pt}
			\includegraphics[width=1.0\textwidth,keepaspectratio]{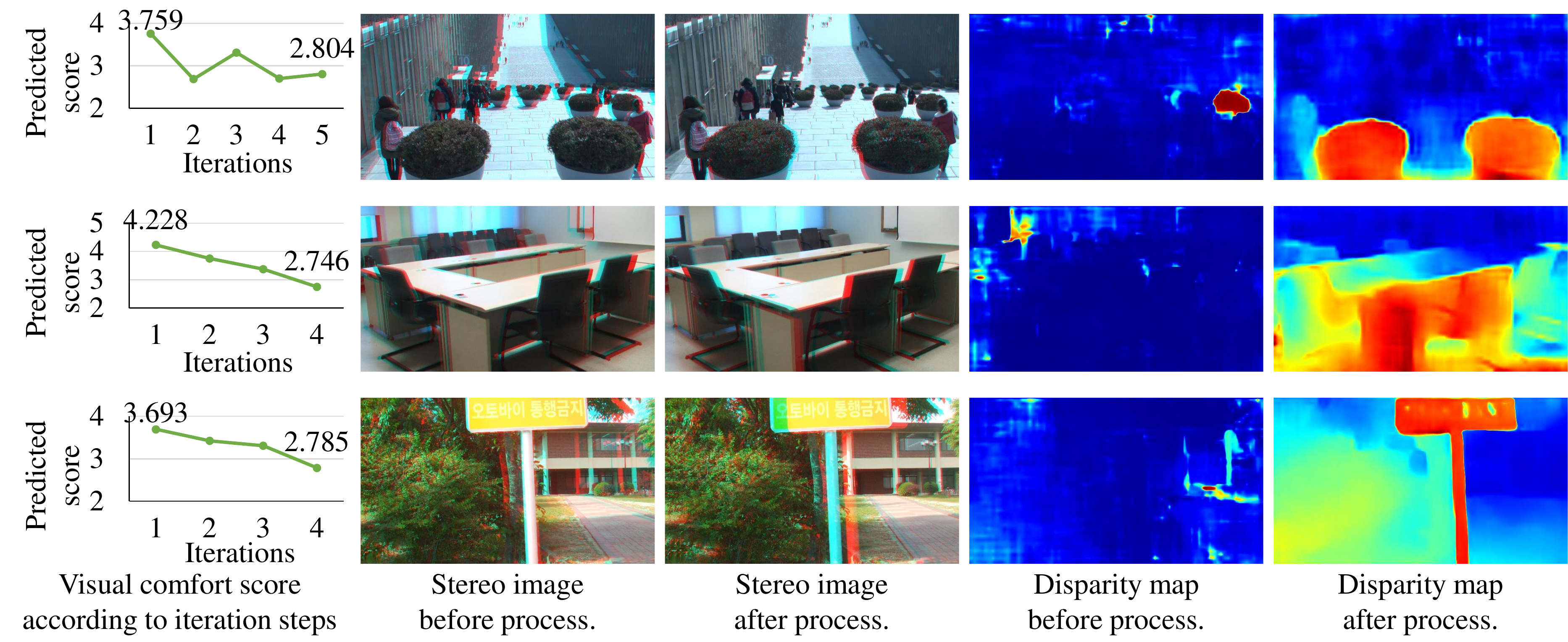}
			\caption{}
			\label{fig3:b}
     \end{subfigure}

\caption{Visual results of our VCA-RL model. (a) Results of uncomfortable stereoscopic images with excessive screen disparities and (b) Results of comfortable stereoscopic images with unnoticeable depths. In case of (a), our VCA-RL model progressively improves the visual comfort level while mitigating excessive disparity magnitude. In case of (b), our model enhances the depth perception in the comfortable range.}
\label{fig:3}
\end{figure*}


\section{Experiments and Results}
\subsection{Experimental Setting}
\textbf{Datasets} In the experiments, IEEE-SA stereo image database \cite{34} was used to train our VCA-RL model. It consists of 800 stereoscopic image pairs with a resolution of 1920$\times$1080 pixels and the corresponding subjective comfort scores. These have 160 different scenes with 5 convergence points. For training and testing, we used 10-fold cross-validation. The IEEE SA stereo image database was randomly divided into 10 subsets. 9 subsets were used for training stage and 1 subset was used for testing stage. 

In testing, to verify the robustness and generalization, we conducted the depth adjustment on additional databases, which are NBU 3D-VCA database \cite{35} and IVY Lab S3D image database for visual discomfort reduction \cite{27}. The NBU 3D-VCA database consists of 200 stereoscopic images (1920$\times$1080) with the associated mean opinion score (MOS) for visual comfort. IVY Lab S3D image database consists of 120 stereoscopic images (1920$\times$1080) captured by 3D digital camera with dual lenses (Fujifilm FinePix 3D W3) and the corresponding MOS values as well. They were used in testing only.

\textbf{Implementation Details} In the training stage, the feature extractor and visual comfort score network were pre-trained end-to-end with Adam optimizer. For Adam optimizer, a learning rate was initialized at $1e-5$. $\beta_1$ and $\beta_2$ were set to 0.9 and 0.999, respectively \cite{36}. Then, we trained our agent network. In our training of deep Q-network with reinforcement learning, we adopted an $\epsilon$-greedy policy. The initial value of $\epsilon$ was 1 and the value decreased until $\epsilon=0.1$ in steps of 0.1. The weights for deep Q-network were initialized with normal distribution \cite{32}. We used an experience replay of 2,000 experiences and a batch size of 256.

\subsection{Visual Results of Iterative Depth Adjustment}

Fig.~\ref{fig3:a} shows examples of uncomfortable stereoscopic images with excessive disparities. In this case, our goal is to reduce the level of visual discomfort and fall the excessive disparities into comfortable range. Our VCA-RL iteratively improved the visual comfort level of given stereoscopic images until the comfort scores reach to comfortable range while avoiding unnoticeable depth information of foreground objects. The agent progressively ameliorated the degree of visual comfort by decreasing the distance between stereo cameras until achieving the target visual comfort score range.

Fig.~\ref{fig3:b} shows examples of very comfortable stereoscopic images with unnoticeable depths. In this case, the proposed method iteratively increased the stereo baseline to improve their disparities for sufficient depth perception. Simultaneously, our agent carefully checked the level of visual comfort to prevent the adjusted disparities from causing extreme visual fatigue. As a result, we could provide visually comfortable stereoscopic images with sufficient depth perception. In Fig.~\ref{fig3:b}, the final comfort score was lower than the comfort score of the original stereoscopic image. However, it is still within the visually comfortable range. In particular, the depth of foreground objects considerably increased by our VCA-RL model.

\subsection{Qualitative Comparisons}

\begin{figure}
\begin{center}
\includegraphics[width=1.0\linewidth] {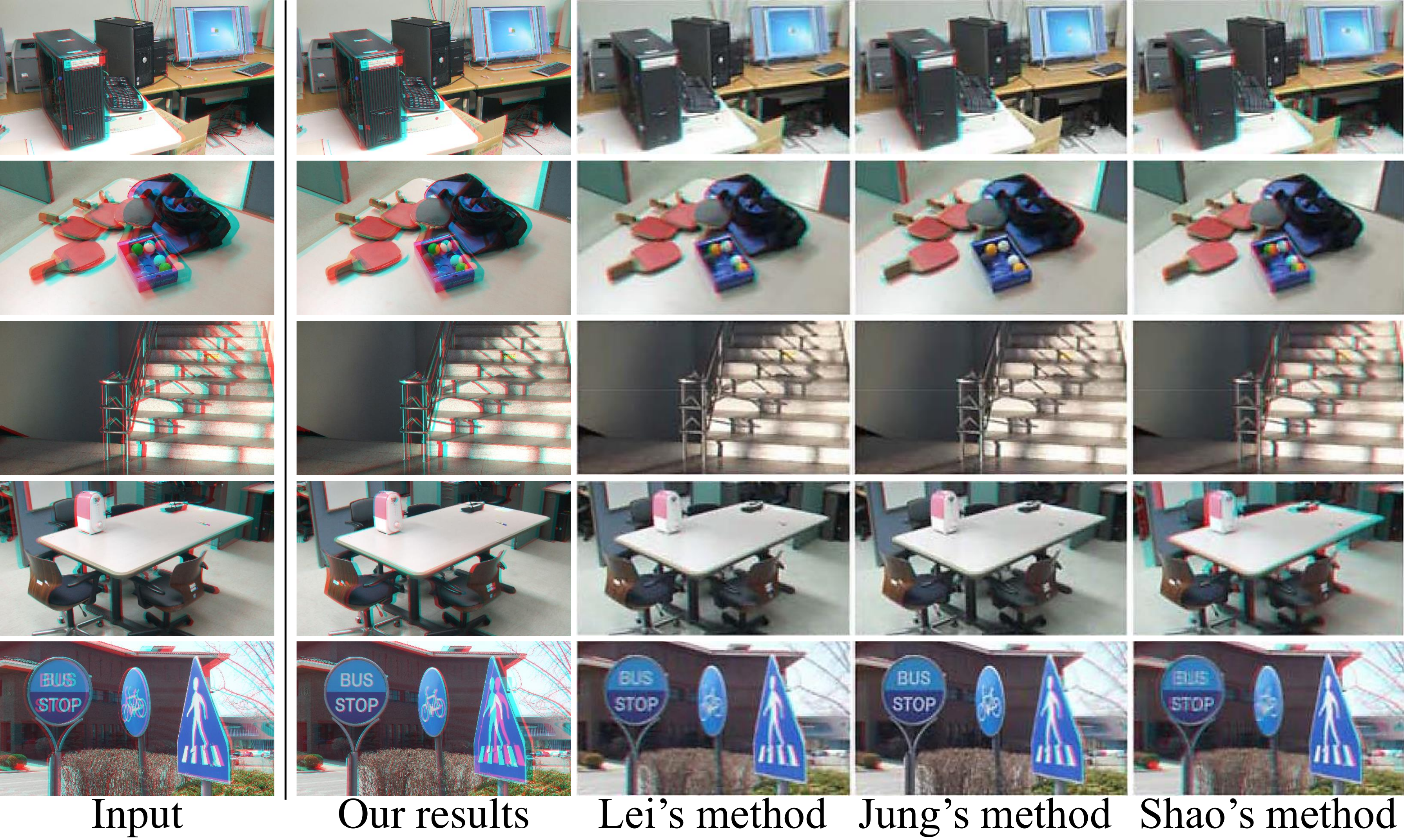}
\end{center}
   \caption{Performance comparisons of depth adjustment in stereoscopic viewing. These are anaglyph images, which can be seen as 3D through red-green glasses. Results of existing methods were taken from each paper.}
\label{fig:4}

\end{figure}

\begin{figure}
\begin{center}
\includegraphics[width=1.0\linewidth] {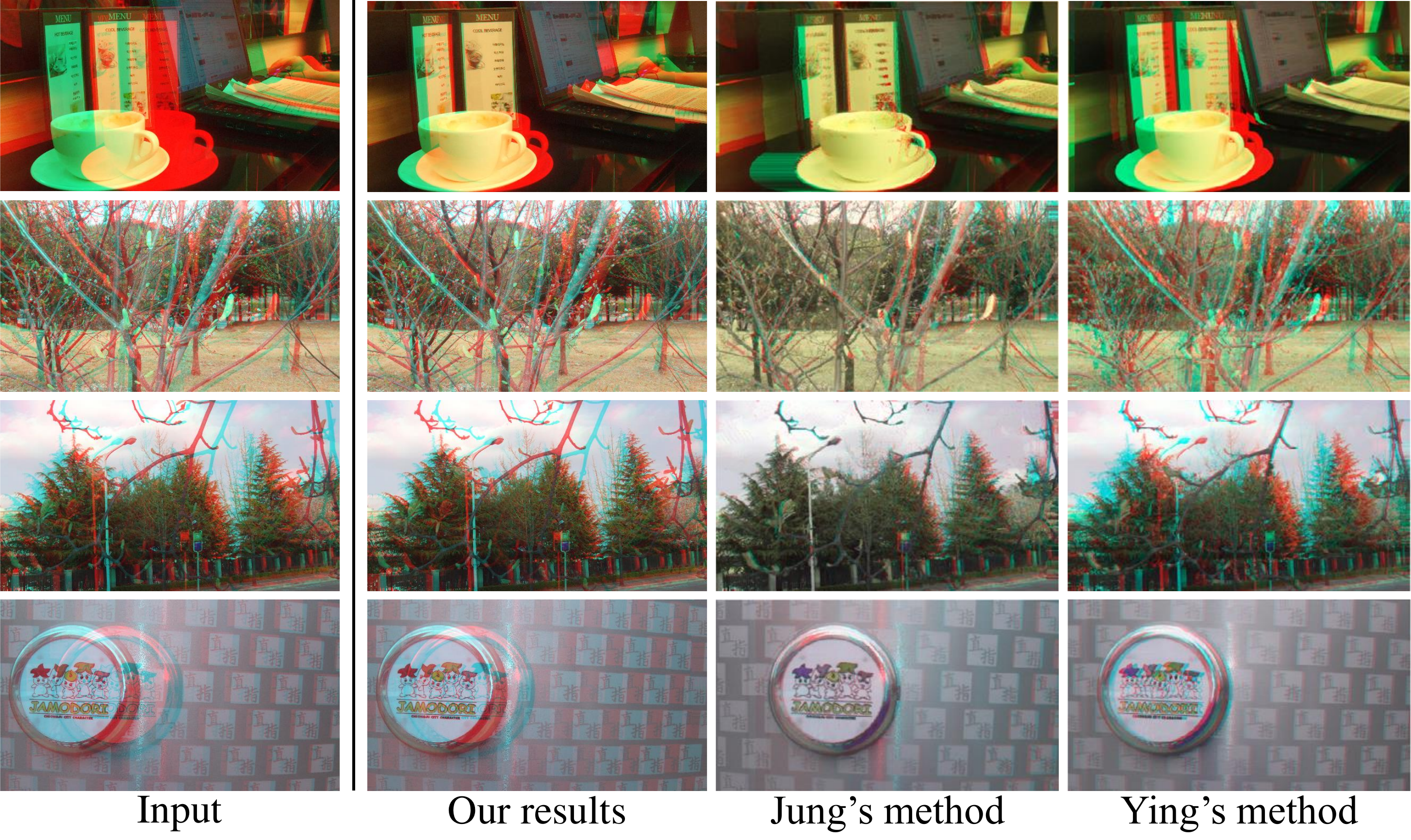}
\end{center}
   \caption{Performance comparisons of depth adjustment in stereoscopic viewing. These are anaglyph images, which can be seen as 3D through red-green glasses. Results of existing methods were taken from each paper.}
\label{fig:5}

\end{figure}


In this section, for performance comparisons of depth adjustment, we visually compared our results with existing depth adjustment methods: Lei’s method \cite{14}, Jung’s method \cite{20}, Shao’s method \cite{23}, and Ying’s method \cite{16}. Fig.~\ref{fig:4} and Fig.~\ref{fig:5} show visual results of our VCA-RL and previous depth adjustment methods \cite{14,16,20,23} for performance comparisons. Note that visual results of previous methods were taken from their papers \cite{14,16,20,23} because their codes and any other results are not available.

In Fig.~\ref{fig:4}, the first input stereoscopic image seems to have sufficient disparities. the inputs in the second and the third rows have large disparities relatively. The disparities of inputs in the fourth and the fifth rows seem to be small. Shao’s method \cite{23} tried to strike a balance between visual comfort and depth perception, compared to \cite{14} and \cite{20}. However, it seemed to fail in the results for the second, the third and the fifth examples. On the other hand, our VCA-RL provided the visual result maintaining the depth for the content in the first row, and visual results mitigating the excessive disparities for the contents in the second and the third rows. For the fourth and the fifth contents, our VCA-RL provided the visual results with increasing disparities.

Fig.~\ref{fig:5} shows visual results of our VCA-RL, Jung's method \cite{20}, and Ying's method \cite{16}. Similar to Fig.~\ref{fig:4}, the result of Jung's method didn't seem large enough to perceive depth. Ying's method \cite{16} provided reliable results for examples in the second and the third rows. However, for the content in the first row, it did not sufficiently reduce the disparities around the menu board. For that in the fourth row, Ying's method \cite{16} also significantly reduced its disparities. On the other hand, our VCA-RL could stably provide reliable visual results for various examples.

\begin{table*}[t!]
  
\centering
\begin{tabular}{c|c|c|c|c|c|c|c|c}
\hline
\hline
\multirow{2}{*}{} & \multicolumn{4}{c|}{Most \textit{uncomfortable} top 10\% stimuli}                      & \multicolumn{4}{c}{Most \textit{comfortable} top 10\% stimuli}                        \\ \cline{2-9} 
                  & VC\_input          & VC\_ours         & DP\_input          & DP\_ours         & VC\_input          & VC\_ours         & DP\_input          & DP\_ours         \\ \hline
mean              & 1.65               & 3.00             & 4.85               & 3.59             & 4.30              & 3.01             & 2.81               & 3.70             \\ \hline
std               & 0.43               & 0.54             & 0.23               & 0.44             & 0.15              & 0.26             & 0.31               & 0.45             \\ \hline
$p$-value           & \multicolumn{2}{c|}{$p$\textless{}0.05} & \multicolumn{2}{c|}{$p$\textless{}0.05} & \multicolumn{2}{c|}{$p$\textless{}0.05} & \multicolumn{2}{c}{$p$\textless{}0.05} \\
\hline
\hline
\end{tabular}
\label{table:1}
\centering
\caption{Statistical results of objective assessment for visual comfort and depth perception on IEEE SA stereo image database}
\end{table*}

\subsection{Quantitative Evaluations}
To verify the effectiveness of our VCA-RL, we objectively measured the levels of visual comfort and the depth perception for most uncomfortable top 10\% and most comfortable top 10\% stereoscopic images on \cite{34}. In this experiment, we employed an objective visual comfort assessment metric (VC score) based on deep visual and disparity feature \cite{4} for visual comfort assessment. To measure the presence of depth, we employed an objective assessment metric (DP score) in \cite{16}. The VC score and the DP score are higher, the comfort level and the perceived depth level are higher. VC and DP scores range from 0 to 5.

The statistical results of the objective assessment on IEEE SA stereo image database \cite{34} are presented in Table 1. For most uncomfortable stimuli, the improvement of overall visual comfort was statistically significant, compared with original input stereoscopic images. The level of depth perception was still good (DP score $>$ 3). For most comfortable stereoscopic images with unnoticeable depth, the improvement of depth perception was statistically significant, compared to the original. More importantly, despite the increase of disparity/depth, the overall visual comfort still remained within comfortable range. Thus, the results revealed that our VCA-RL could provide a significantly meaningful improvement in terms of both visual comfort and depth perception (see Table S1 and S2 in our supplementary material for statistical results on NBU \cite{35} and IVY databases \cite{27}).

\begin{table}[t]
\centering
	\resizebox{\columnwidth}{!}{
\begin{tabular}{c|c|c|c|c}
\hline
\hline
\multirow{2}{*}{}                                     & \multicolumn{2}{c|}{For \textit{uncomfortable} stimuli}                                                                           & \multicolumn{2}{c}{For \textit{comfortable} stimuli}                                                                             \\ \cline{2-5} 
                                                      & \begin{tabular}[c]{@{}c@{}}before\\ processing\end{tabular} & \begin{tabular}[c]{@{}c@{}}after\\ processing\end{tabular} & \begin{tabular}[c]{@{}c@{}}before\\ processing\end{tabular} & \begin{tabular}[c]{@{}c@{}}after\\ processing\end{tabular} \\ \hline
\begin{tabular}[c]{@{}c@{}}mean of\\ MOS\end{tabular} & 2.81                                                        & 3.67                                                       & 3.7                                                         & 3.26                                                       \\ \hline
std                                                   & 0.38                                                        & 0.18                                                       & 0.17                                                        & 0.36                                                       \\ \hline
\textit{p}-value                                               & \multicolumn{2}{c|}{\textit{p}\textless{}0.05}                                                                                    & \multicolumn{2}{c}{\textit{p}\textless{}0.05}                                                                                    \\
\hline
\hline
\end{tabular}
\label{table:2}
\centering
}
  \caption{Statistical results of subjective assessment on IVY S3D database}
\end{table}

\subsection{User Study}
Furthermore, we conducted a set of subjective assessment experiments to investigate users' visual comfort rating and viewing preference. A half-mirror type stereoscopic 3D monitor was used to display the stereoscopic images. A total of 16 subjects participated in the experiment. We randomly selected 15 uncomfortable stimuli among stereoscopic images with $\hat{s}_{VC}$ less than 3 and 15 comfortable stimuli among stereoscopic images with $\hat{s}_{VC}$ higher than 3 on IVY Lab S3D image database \cite{27}.

To measure the degree of visual comfort, a modified version of the single stimulus (SS) was used with a five point grading scale \cite{37}. During the experiment, original images (‘before processing’) and our results (‘after processing’) were randomly presented to the subjects (i.e., a total of 60 stimuli). For the viewing preference test, subjects were asked to answer the following question: “Which one do you prefer to see in considering all quality aspects of the viewing experience of stereoscopic images?”\cite{13}. For more details, please see the section for subjective assessment environment and procedure in our supplementary material.

Table 2 shows the statistical analysis of subjective assessment results. For uncomfortable stimuli, the mean of MOS value after processing increased statistically significantly than the mean of MOS before processing. The mean of difference MOS was $+$0.86 in range of [1, 5] (i.e., 21.5\% improvement). For comfortable stimuli, the mean of MOS values after processing decreased statistically significantly than the mean of MOS before processing. After enhancing the depth perception, the mean of MOS after processing (i.e., 3.26) remained within comfortable range. These results demonstrated that our agent carefully adjusted depths to provide better visual comfort for uncomfortable data and to preserve visual comfort for comfortable data, respectively.

\begin{figure}[t!]
\begin{center}
\includegraphics[width=0.95\linewidth] {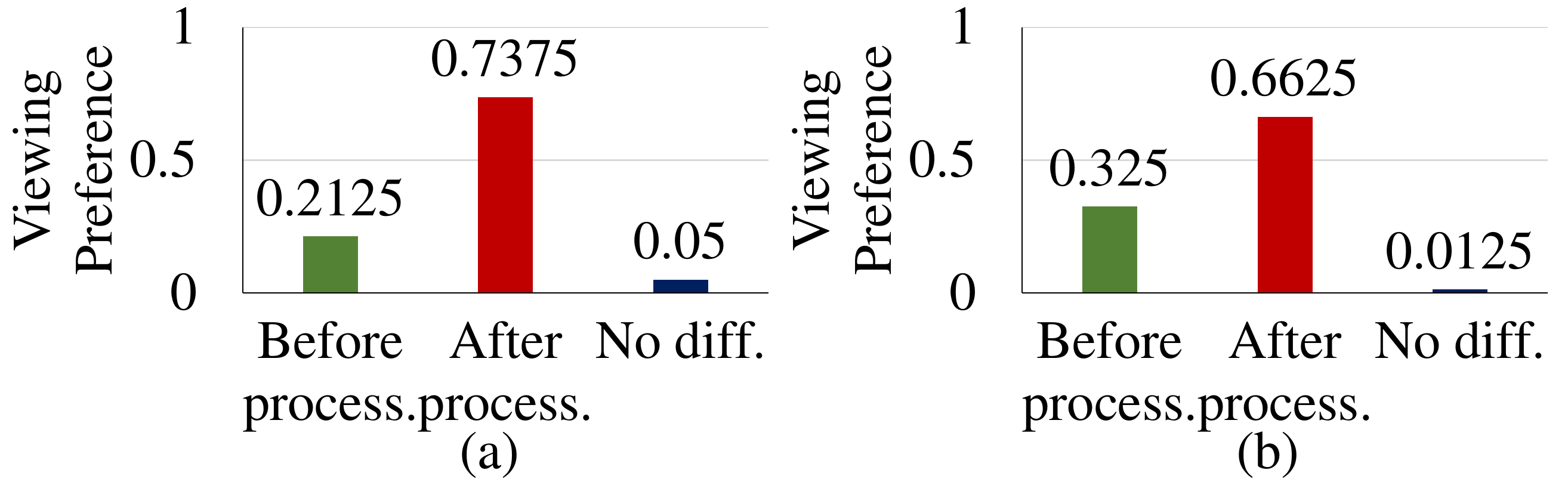}
\end{center}
   \caption{Subjective assessment result for viewing preference. (a) For uncomfortable stereoscopic images. (b) For comfortable stereoscopic images. ‘No diff.’ means there is no difference between before and after processing.}
\label{fig:6}

\end{figure}

Fig.~\ref{fig:6} shows the results of the viewing preference. The viewing preference of ‘after processing’ was much better than that of ‘before processing’. In summary, the result demonstrated that the proposed method had a positive effect on the overall viewing experience of stereoscopic images by carefully increasing or decreasing their depths.

\section{Conclusion}
In this paper, we proposed a novel reinforcement learning-based approach considering visual comfort of stereoscopic images for depth adjustment named VCA-RL. With the deep reinforcement learning strategy, we explicitly modeled a human professional’s depth adjustment process. In particular, to take into account perceptual aspects in stereoscopic viewing, we designed a novel visual comfort aware reward function to train our agent to learn the perceptual characteristics of stereoscopic viewing. Therefore, our VCA-RL could sequentially estimate proper depth adjustment steps via DIBR process. With extensive qualitative and quantitative experiments on various S3D databases, our VCA-RL model showed its effectiveness and superiority for depth adjustment. In addition, the results of user study showed that our VCA-RL could be feasible for current S3D displays.

\section{Acknowledgements}
This work was partly supported by IITP grant (No. 2017-0-00780), IITP grant (No. 2017-0-01779), and BK 21 Plus project. M. Park is now in ETRI, Korea.


\bibliography{refs_aaai}

\end{document}